# Multi-stage robust nonlinear model predictive control of a lower-limb exoskeleton robot

**Alireza Aliyari**  **, and Gholamreza Vossoughi**

Department of Mechanical engineering, Sharif university of technology, Tehran, Iran.

Corresponding author: **Alireza Aliyari**  (e-mail: alireza.aliyari@alum.sharif.edu)

## Abstract

The use of exoskeleton robots is increasing due to the rising number of musculoskeletal injuries. However, their effectiveness depends heavily on the design of control systems. Designing robust controllers is challenging due to uncertainties in human-robot systems. Among various control strategies, Model Predictive Control (MPC) is a powerful approach due to its ability to handle constraints and optimize performance. Previous studies used linearization-based methods to implement robust MPC on exoskeletons, but these can degrade performance due to nonlinearities in the robot's dynamics. To address this gap, this paper proposes a Robust Nonlinear Model Predictive Control (RNMPC) method, called multi-stage NMPC, to control a two-degree-of-freedom exoskeleton by solving a nonlinear optimization problem. This method uses multiple scenarios to model system uncertainties. The study focuses on minimizing human-robot interaction forces during the swing phase, particularly when the robot carries unknown loads. Simulations and experimental tests show the proposed method significantly improves robustness, outperforming non-robust NMPC. It achieves lower tracking errors and interaction forces under various uncertainties. For instance, when a 2 kg unknown payload combined with external disturbances, the RMS values of thigh and shank interaction forces for multi-stage NMPC are 77% and 94% lower, respectively, compared to non-robust NMPC.

## 1. Introduction

The use of exoskeleton robots is increasing due to the rising number of musculoskeletal injuries (Theurel J & Desbrosses 2019). Power augmentation is one of the applications of these robots in which the user specifies the motion trajectory and can maintain balance and avoid obstacles. The robot's task is to track human motion and reduce the user's physical effort. Power augmenting exoskeletons can be used to minimize human-robot interaction forces or generate interaction torques in the same direction as human joint torques. During the swing phase of human normal gait, negligible muscle effort is required. At this phase, exoskeleton assistance is unnecessary and could cause discomfort. The goal is to preserve natural gait without interference. To achieve this, the robot should precisely track human movement and minimize human-robot interaction forces during this phase. To achieve these objectives effectively, the control system of the exoskeleton must be precisely designed. Accurate control of exoskeleton robots is critical to ensure smooth motion tracking and enhance user comfort during the swing phase.

Due to the nonlinearity of the robot's dynamics, system uncertainties, and direct interaction of exoskeleton robots with humans, a nonlinear robust and accurate control system must be designed for these robots (Aliman et al. 2021; Tahamipour et al. 2021; Guo et al. 2015). Extensive research has been done with the aim of designing robust controllers for exoskeleton robots (Kardan & Akbarzadeh 2017; Brahmi et al. 2018; Mefoued & Belkhiat 2019; Narayan & Dwivedy 2021; Mokhtari et al. 2021; Aliman et al. 2022; Rose et al. 2022). Model predictive control (MPC) considers system constraints in control design and is capable of controlling complex systems (Mayne et al. 2000; Prasetyo et al. 2017). These characteristics make MPC a suitable controller for exoskeleton robots (Jammeli et al. 2021). Teramae et al. (2017) used an implicit non-robust MPC to control a 1-DOF upper-limb exoskeleton robot. Wang et al. (2011) presented a linear MPC by linearizing the robot's dynamics around its set point to control a lower-limb exoskeleton robot during the swing phase of normal gait. Their control system required predefined parameters such as stride length and walking speed. Hu et al. (2013) used feedback linearization combined with linear MPC for a 2-DOF upper-body exoskeleton robot. Their proposed control scheme eliminated nonlinear terms of the robot's dynamics by a feedback linearization controller, after which an explicit MPC was used to control the remaining linear terms. Cao & Huang (2020) implemented a nonlinear model predictive control (NMPC) which was optimized by the gradient descent method to control an exoskeleton robot equipped with pneumatic actuators. An echo state network was employed to predict the robot's dynamics over the prediction horizon. Tahamipour et al. (2021) linearized the dynamic model of the exoskeleton robot at each time step to implement a linear robust MPC which was optimized by the active set method.

Based on the existing literature, only a limited number of researchers have employed Robust Model Predictive Control (MPC) for controlling exoskeleton robots. These studies have primarily relied on linearization-based methods to implement Robust MPC. This can adversely affect the performance of the control system due to the inherent nonlinearities of robot's dynamic model. This work addresses this gap by proposing a Robust Nonlinear Model Predictive Control (RNMPC) approach named the multi-stage NMPC method. It controls a 2-DOF exoskeleton robot by solving a nonlinear optimization problem. The primary objectives of this work are to track human motion and minimize human-robot interaction forces during the swing phase of normal gait, particularly under the condition of an unknown external load carried by the robot, external disturbances, and dynamic model uncertainties.

The rest of this paper is organized as follows. The proposed models for the human and robot during the swing phase and the model used for generating human-robot interaction forces in the simulation environment are presented in section 2. The proposed NMPC approaches are presented in section 3. The simulation results of the proposed controllers are reported in section 4. Finally, some concluding remarks are provided in section 5.

## 2. Human and exoskeleton robot models

Dynamic models for the human and robot in the sagittal plane are presented in this section. For simplicity, leg movements in other planes are ignored, and each limb of the human and robot is modeled as a double pendulum (Fig. 1). The models consist of shank and thigh links, with two actuators are employed for the hip and knee joints of the robot. The human and robot are connected through two straps placed on the shank and thigh. Human-robot interaction forces are transmitted through these straps.

**Fig. 1.** Dynamic model of human (red) and the exoskeleton (green).

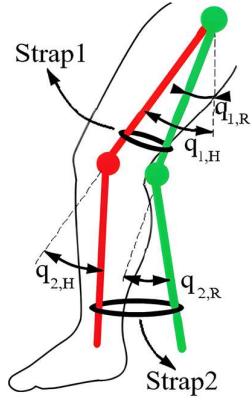

Using the Euler-Lagrange equations, the dynamic models of the human and robot can be described by eqs. 1 and 2

(1) $\quad M_R(q_R)\ddot{q}_R + C_R(q_R,\dot{q}_R)\dot{q}_R + G_R(q_R) + k_{f1}\dot{q}_R + k_{f2}\,sign(\dot{q}_R) + D = T_R + T_{int}$

(2) $\quad M_H(q_H)\ddot{q}_H + C_H(q_H,\dot{q}_H)\dot{q}_H + G_H(q_H) = T_H - T_{int}$

Where $q_R = [q_{R,Hip}, q_{R,Knee}]^T$ and $q_H = [q_{H,Hip}, q_{H,Knee}]^T$ are the robot and human joint angle vectors, $\dot{q}_R = [\dot{q}_{R,Hip}, \dot{q}_{R,Knee}]^T$ and $\dot{q}_H = [\dot{q}_{H,Hip}, \dot{q}_{H,Knee}]^T$ are the robot and human joint angualr velocity vectors, $\ddot{q}_R = [\ddot{q}_{R,Hip}, \ddot{q}_{R,Knee}]^T$ and $\ddot{q}_H = [\ddot{q}_{H,Hip}, \ddot{q}_{H,Knee}]^T$ are the robot and human joint angular acceleration vectors, $T_R = [T_{R,Hip}, T_{R,Knee}]^T$ and $T_H = [T_{H,Hip}, T_{H,Knee}]^T$ are the robot and human joint torque vectors, respectively. The joint torques of the robot are control inputs calculated by the control system. $M_R$ and $M_H$ are the mass matrices of the robot and human, $C_R$ and $C_H$ are Coriolis matrices of the robot and human, $G_R$ and $G_H$ are the gravitational vectors of the robot and human, and $T_{int} = [T_{int,Hip}, T_{int,Knee}]^T$ represents human-robot interaction torque vector, which is obtained by multiplying the interaction forces by the lever arm. Additionally $k_{f1}$ and $k_{f2}$ are the viscous and coulomb friction coefficients, respectively, and $D$ denotes the total disturbance of the system.

Interaction forces couple the robot's dynamic equations with those of the human model. To avoid interfering with human movement, it is essential to minimize these forces. By increasing the angular position or velocity difference between human and robot joints, the interaction forces will increase. While force sensors can measure these forces experimentally, they are inaccessible in the simulation environment. Therefore, human-robot interaction is modeled using a spring-damper system. Equations 3 and 4 are used to generate these forces:

(3) $F_{int,1} = k_s L_{s,1}(q_{H,Hip} - q_{R,Hip}) + c_s L_{s,1}(\dot{q}_{H,Hip} - \dot{q}_{R,Hip})$

(4) $F_{int,2} = k_s [L_{s,1}(q_{H,Hip} - q_{R,Hip}) + L_{s,2}(q_{H,Knee} - q_{R,Knee})] + c_s [L_{s,1}(\dot{q}_{H,Hip} - \dot{q}_{R,Hip}) + L_{s,2}(\dot{q}_{H,Knee} - \dot{q}_{R,Knee})]$

Where $F_{int,1}$ and $F_{int,2}$ denote the interaction forces of the hip and knee joints, respectively. $L_{s,1}$ is the distance of strap 1 from the hip joint, and $L_{s,2}$ is the distance of strap 2 from the knee joint. Additionally, $k_s$ and $c_s$ denote the stiffness and damping coefficients of the modeled spring-damper system.

The control system of the robot requires the angular position and velocity of human joints as the desired outputs. However, since it is not possible to measure these values, they can be approximated using the interaction forces according to eqs. 5 and 6:

(5) $\hat{q}_H = q_R + K_1 F_{int}$

(6) $\hat{\dot{q}}_H = \dot{q}_R + K_2 F_{int}$

Where $\hat{q}_H = [\hat{q}_{H,Hip}, \hat{q}_{H,Knee}]^T$ and $\hat{\dot{q}}_H = [\hat{\dot{q}}_{H,Hip}, \hat{\dot{q}}_{H,Knee}]^T$ are the estimated vectors of human joint angular positions and velocities, respectively; and are considered as the desired outputs of the robot control system. $K_1$ and $K_2$ are diagonal matrices with non-negative elements.

## 3. Control design

This section discusses the proposed control system. First, a non-robust NMPC is developed. Then, a multi-stage NMPC is designed as a robust NMPC to improve the controller's performance against uncertainties.

### 3.1. Non-robust NMPC

Model predictive control is often implemented in discrete-time settings. At each time step, MPC calculates the sequence of control inputs over the prediction horizon by optimizing a predefined cost function. Then, the first element of the calculated control inputs is applied to the system. The calculations are then repeated for the next time step and the new prediction horizon. In order to use MPC, the state variables must be predicted over the prediction horizon, which is achieved using the dynamic model, current state variables, and system constraints. The continuous-time model of the robot is discretized using the 4$^{th}$-order Runge-Kutta method, and the optimization problem is formulated as eq. 7.

(7) $\min_{\Delta T_R} J = \sum_{j=1}^{N_p} ||\hat{q}_R(k+j|k) - \hat{q}_H(k+j)||^2 + r_d \sum_{j=1}^{N_p} ||\hat{\dot{q}}_R(k+j|k) - \hat{\dot{q}}_H(k+j)||^2 + r_t \sum_{j=0}^{N_c} ||\Delta T_R(k+j)||^2$

subject to eqs. 8 and 9.

(8) $T_{R,min} \leq T_R(k+j) \leq T_{R,max}$

(9) $\Delta T_{R,min} \leq \Delta T_R(k+j) \leq \Delta T_{R,max}$

Where $\hat{q}_R(k+j|k)$ and $\hat{\dot{q}}_R(k+j|k)$ denote predicted angular positions and velocities over the prediction horizon, which are calculated using the dynamic equations of the robot. $\Delta T_R(k+j) = [\Delta T_{R,Hip}(k+j), \Delta T_{R,Hip}]^T$ is the j-th element of the control increment sequence. $N_p$, $N_c$, $r_d$ and $r_t$ are the prediction horizon, control horizon, weighting coefficient of angular velocity tracking and weighting coefficient of control effort, respectively. The first and second terms in eq. 7 are used to reduce the tracking error of angular positions and velocities, respectively, while the third term is used to reduce the control increments. Equations 8 and 9 represent the constraints of the optimization problem, which aim to limit the magnitude of the control inputs and control increments, respectively.

Since the dynamic equations of the robot contain non-linear terms, the optimization problem is not necessarily convex and an analytical solution cannot be obtained. Therefore, the optimization problem must be solved numerically. Sequential quadratic programming (SQP) (Gill et al. 2005) is used to solve the constrained nonlinear optimization problem. SQP is a fast and accurate method for calculating the optimal control inputs.

## 3.2. Robust NMPC

Designing a robust MPC scheme is a significant challenge (Lucia & Karg 2018). Tube-based MPC (Mayne et al. 2011), Min-Max (Bemporad et al. 2003) and multi-stage NMPC (Bernardini & Bemporad 2009; Lucia et al. 2012; Lucia et al. 2013) are some of the robust MPC algorithms. In this paper, multi-stage NMPC (Lucia et al. 2013; Jia & You 2021), powerful control scheme, is used to control the exoskeleton robot. In this method, different values for the system's uncertainties are selected within a specific interval. By developing a scenario tree, different scenarios are considered based on the selected uncertainties, and the state variables and the sequence of control inputs over the prediction horizon are calculated using the dynamic equations of each scenario. Then, the probability of occurrence of each scenario is determined using parallel Extended Kalman filters (EKF). Another advantage of using EKFs is their ability to estimate external disturbances, which significantly helps in achieving a robust controller. The proposed scenario tree is shown in Fig. 2. An arbitrary uncertainty is considered for each branch of the scenario tree. By comparing the predicted outputs of each scenario with the actual outputs of the system, the probability of occurrence of the i-th scenario is calculated using EKF based on eqs. 10 and 11.

$$(10) \quad \acute{\mu}_i(k) = \frac{\frac{\exp(-0.5|c_1 v_i(k) \bar{S}_i(k)^{-1} v_i(k)^T|)}{\sqrt{(2\pi)^{\dim(y(k))} |c_1 \bar{S}_i(k)|}} \mu_i(k-1)}{\sum_{j=1}^{N} \frac{\exp(-0.5|c_1 v_j(k) \bar{S}_j(k)^{-1} v_j(k)^T|)}{\sqrt{(2\pi)^{\dim(y(k))} |c_1 \bar{S}_j(k)|}} \mu_j(k-1)}$$

$$(11) \quad \mu_i(k) = \max\{\acute{\mu}_i(k), 10^{-4}\}$$

Where $\mu_i(k)$ denotes the probability of occurrence of the i-th scenario at time step $k$. $y(k) = [q_R(k)^T, \dot{q}_R(k)^T]^T$ and $v_i(k) = \left[\left(q_R(k) - \hat{q}_{R,i}(k|k-1)\right)^T, \left(\dot{q}_R(k) - \hat{\dot{q}}_{R,i}(k|k-1)\right)^T\right]^T$ are, respectively, the actual output vector of the system and the estimation error of the i-th scenario at time step $k$. The predicted output of each scenario is the sum of the output calculated by dynamic equations of that scenario and the corresponding estimation error at the previous time step. $\bar{S}_i$ is one of the parameters of the EKF, and its update method is described in (Subramanian et al. 2014). $N$ represents the number of scenarios, and $c_1$ is a constant, that controls the speed of convergence.

The multi-stage optimization problem is formulated as eq. 12.

$$(12) \quad \min_{\Delta T_R} J = \sum_{i=1}^{N} \left[ \mu_i(k) \left( \sum_{j=1}^{N_p} \left\| \hat{q}_{R,i}(k+j|k) - \hat{q}_H(k+j) \right\|^2 + r_d \sum_{j=1}^{N_p} \left\| \hat{\dot{q}}_{R,i}(k+j|k) - \hat{\dot{q}}_H(k+j) \right\|^2 + r_t \sum_{j=0}^{N_c} \left\| \Delta T_{R,i}(k+j) \right\|^2 \right) \right]$$

According to eq. 12, the number of optimization variables in multi-stage NMPC is greater than in non-robust NMPC, which leads to an exponential increase in the processing time. However, due to the structure of the proposed scenario tree and the independence of the scenarios, the optimization problem is solved separately for each scenario. This strategy reduces the processing time. Therefore, $N$ optimization problem with the cost function given in eq. 7 are solved using SQP, and the control input is calculated according to eq. 13.

$$(13) \quad T_R(k) = T_R(k-1) + \sum_{i=1}^{N} \mu_i(k) \Delta T_{R,i}(k)$$

**Fig. 2.** The proposed scenario tree.

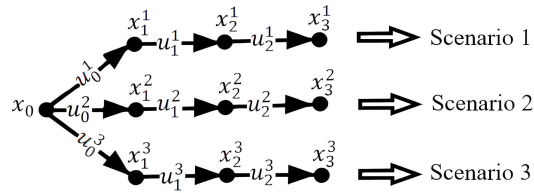

4## 4. Results

In this section, simulation and experimental results are presented to illustrate the performance of multi-stage NMPC for controlling the exoskeleton robot and to compare it with non-robust NMPC. A concentrated mass of 0 to 2 kg, as an unknown external load carried by the robot, is placed in the middle of the robot's shank link. Its magnitude remains constant during each simulation. Three different scenarios with uncertainties of 0, 1 and 2 kg are considered in the scenario tree. The probability of occurrence of these scenarios is indicated by $\mu_1$, $\mu_2$ and $\mu_3$, respectively. The maximum value of torques and their rate of change are set to 30 N.m and 1000 N.m/s, respectively. The time step for both the simulation and experimental tests is set to 10 ms.

### 4.1. Simulation

In the simulation environment, periodic trajectories with variable frequency and amplitude are provided as the desired signal to the human model, and a feedback linearization control is used to simulate the behavior of the human brain. The robot's controller does not have any data about the human model and it's kinematic variables and only human-robot interaction forces are provided for the robot's controller. The model and control parameters are presented in Table 1. In addition to a constant mass of 0-2 Kg, an uncertainty of $m_{dis} = 0.05\sin(t) + 0.01\sin(100t + \frac{\pi}{4})$ is introduced for the masses of the shank and thigh links of the robot. Meanwhile, a term of $0.1\sin(t) + 0.05\sin(100t + \frac{\pi}{2})$ for both hip and knee joints is considered as the remaining uncertainties and unmodeled dynamics, which appear in matrix $D$. Furthermore, terms of $10^{-4}\sin(100t)$ and $10^{-3}\sin(100t)$ are considered as measurement noise for the velocity and force sensors, respectively. To reduce this noise, the average of the five most recent measurements of velocities and interaction forces is introduced into the model, which leads to an undesired delay in the control system. Additionally, to evaluate the control system's robustness to human physiological changes, the human mass was increased linearly from 60 kg (initial condition) to 85 kg (final condition) during the simulation period. The effects of all these uncertainties are evaluated in the simulation environment.

**Table 1.** Model and control parameters.

| Model parameters | | Control parameters | |
| --- | --- | --- | --- |
| Parameter | Value | Parameter | Value |
| $k_{f1}$(N.m.s) | 0.000899 | $N_p$ | 3 |
| $k_{f2}$(N.m) | 0.05048 | $N_c$ | 3 |
| $k_s$(N/m) | 937.5 | $N$ | 3 |
| $c_s$(N.s/m) | 93.7 | $r_d$ | 0.1 |
| $L_{s,1}$(m) | 0.28 | $r_t$ | 0.000001 |
| $L_{s,2}$(m) | 0.16 | $c_1$ | 100 |
| $K_1$(1/N) | $0.005 \times I_{2 \times 2}$ | | |
| $K_1$(1/N.s) | $0.05 \times I_{2 \times 2}$ | | |

Simulation results of controlling the exoskeleton robot using multi-stage NMPC for the case where the magnitude of the unknown mass is 2 kg, are shown in Fig. 3 and Fig. 4. Fig. 3 shows the hip and knee joints angular position and velocity tracking and control input calculated by the proposed controller. Fig. 4 shows the tracking errors of the joints and interaction forces obtained by multi-stage NMPC (MSNMPC) and non-robust NMPC, and also, the probability of occurrence of scenarios in multi-stage NMPC. The results demonstrate that multi-stage NMPC performs better than non-robust NMPC. Since the control time step is limited to 10 ms for real-time operation, the computation time for control inputs must be less than this limit to ensure system stability. Fig. 5 demonstrates the computation time required for calculating the control signal using a Ryzen 4650G CPU as the robot's hardware controller. The average calculation time in each time step is 3.7 ms. The results confirm that this hardware achieves real-time performance. The presented graph corresponds to a sequential computation process. For weaker hardware to meet real-time requirements, parallel computation can be employed due to the independence of scenarios in the scenario tree. The multi-stage NMPC formulation requires solving three parallel scenario optimizations per time step, while the non-robust NMPC computes only a single optimization. Consequently, the non-robust controller achieves significantly faster computation times (average 1.05 ms vs. 3.7 ms).

**Fig. 3.** Multi-stage NMPC joints angle (left graphs), angular velocity (middle graphs) and control inputs (Right graphs) for 2Kg load.

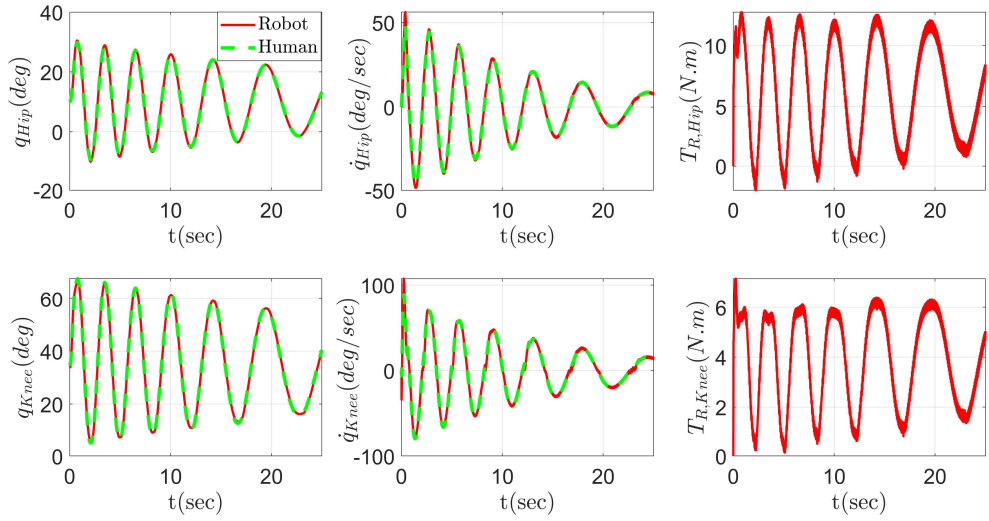

**Fig. 4.** NMPC and Multi-Stage NMPC tracking error of the angles (left graph), NMPC and Multi-Stage NMPC interaction forces (middle graph), and the probability of occurrence of the scenarios in Multi-Stage NMPC for 2Kg load (right graph).

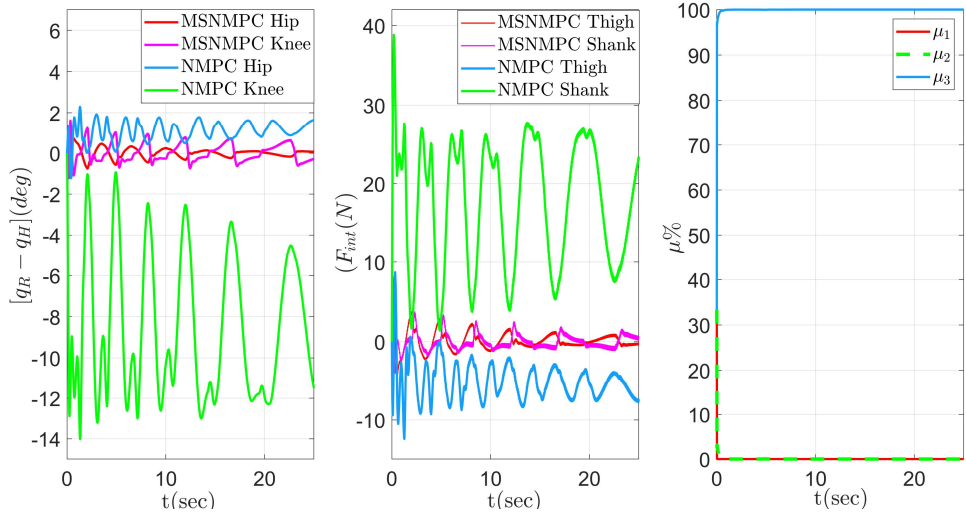

**Fig. 5.** The computation time of the controller for a 2-DOF robot using a Ryzen 4650G CPU.

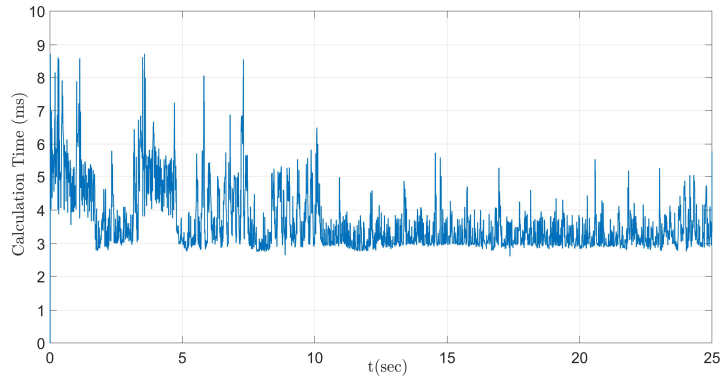

Increasing parameter $r_t$ in eq. 12 yields smoother control signals at the cost of higher interaction forces. Similarly, raising $r_d$, $K_1$ and $K_2$ (eqs. 12, 5, and 6) amplifies system fluctuations, though setting these parameters too low may induce instability. While expanding the control horizon or scenario count improves performance, it drastically increases computation time—exceeding the real-time capabilities of our Ryzen 4650G hardware for large horizons.

The results obtained for the case with no uncertainty are indicated in Fig. 6. According to these results, multi-stage NMPC outperforms non-robust NMPC due to its capability to estimate external disturbances using the EKF. The Root Mean Square (RMS) of interaction forces ($RMS_{F_{int}}$), the maximum value of tracking error of the angles ($\delta_{max}$) obtained by multi-stage NMPC and non-robust NMPC, and the average probability of occurrence of scenarios ($\bar{\mu}$) in multi-stage NMPC for various uncertainties are presented in Table 2. These results further confirm the superior performance of multi-stage NMPC. These results demonstrate that interaction forces under non-robust NMPC increase significantly with payload, while multi-stage NMPC maintains near-constant forces due to its inherent robustness. Under 2 kg payload conditions, non-robust NMPC exhibited dramatic force amplification in relative to 0 kg payload: +622% (thigh) and +1272% (shank). In striking contrast, multi-stage NMPC demonstrated near-constant force profiles, showing only +26% (thigh) variation and a -6.8% (shank) reduction, confirming its superior disturbance rejection capability.

**Fig. 6.** NMPC and Multi-Stage NMPC tracking error of the angles (top graph), Multi-Stage NMPC interaction forces (middle graph) and the probability of occurrence of the scenarios in Multi-Stage NMPC (bottom graph) in the case where there is no load.

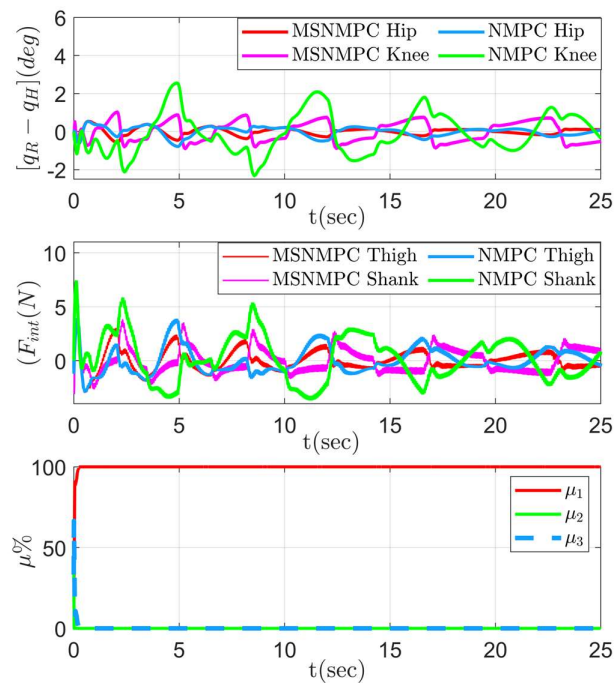

**Table 2.** NMPC and Multi-Stage NMPC RMS interaction forces, NMPC and Multi-Stage NMPC maximum value of tracking error of the angles and the average probability of occurrence of the scenarios in Multi-Stage NMPC for different varieties of uncertainties.

| The magnitude of the uncertain mass (Kg) | $RMS_{F_{int}}$ (N) | | | | $\delta_{max}$ (deg) | | | | $\bar{\mu}$ % | | |
|---|---|---|---|---|---|---|---|---|---|---|---|
| | Multi-stage NMPC | | NMPC | | Multi-stage NMPC | | NMPC | | Multi-stage NMPC | | |
| | Thigh | Shank | Thigh | Shank | Hip | Knee | Hip | Knee | $\bar{\mu}_1$ | $\bar{\mu}_2$ | $\bar{\mu}_3$ |
| 0.00 | 0.98 | 1.18 | 0.77 | 1.40 | 0.70 | 1.03 | 0.52 | 1.68 | 99.99 | 0.00 | 0.01 |
| 0.25 | 0.99 | 1.24 | 1.08 | 3.25 | 0.75 | 1.12 | 0.47 | 2.87 | 99.56 | 4.30 | 0.01 |
| 0.50 | 1.01 | 1.38 | 1.81 | 5.78 | 0.81 | 1.17 | 0.77 | 4.70 | 57.56 | 41.12 | 1.32 |
| 0.75 | 1.35 | 1.08 | 2.55 | 8.27 | 1.14 | 1.78 | 1.06 | 6.54 | 10.03 | 79.78 | 10.19 |
| 1.00 | 1.49 | 1.07 | 3.27 | 10.67 | 1.21 | 1.84 | 1.35 | 8.33 | 0.39 | 81.05 | 18.56 |
| 1.25 | 1.30 | 1.18 | 3.95 | 12.96 | 1.35 | 1.99 | 1.64 | 10.00 | 0.27 | 83.10 | 16.64 |
| 1.50 | 1.39 | 1.15 | 4.58 | 15.14 | 1.44 | 2.16 | 1.91 | 11.54 | 0.22 | 7.70 | 92.09 |
| 1.75 | 1.24 | 1.06 | 5.11 | 17.23 | 1.02 | 1.22 | 2.14 | 12.89 | 0.03 | 0.90 | 99.08 |
| 2.00 | 1.24 | 1.10 | 5.56 | 19.22 | 1.17 | 1.55 | 2.29 | 14.01 | 0.02 | 0.02 | 99.96 |

The results demonstrate the effectiveness of multi-stage NMPC in robot control and its superior performance compared to non-robust NMPC when handling various types of external disturbances. To validate these results, experimental tests are necessary. However, since the experiments are conducted on a 1-DOF shank model, corresponding simulations of the 1-DOF robot must first be performed to ensure consistency between simulation and experimental outcomes. The simulation conditions remain similar to those previously defined, with two exceptions: The human mass is set to 60 kg (representing the user's mass), and the desired trajectory of the human follows the same motion profile as in the experimental tests (section 4.2). The simulation results for the case where the magnitude of the unknown mass is 2 kg, are shown in Fig. 7 and Fig. 8. Fig. 7 shows the knee joint's angular position and velocity tracking and control input calculated by the proposed controller. Fig. 8 shows the tracking errors of the knee joint and interaction forces obtained by multi-stage NMPC (MSNMPC) and non-robust NMPC, and also, the probability of occurrence of scenarios in multi-stage NMPC. The 1-DOF simulations demonstrate the functionality of the multi-stage NMPC for a 1-DOF robot, similar to the 2-DOF case. The results indicate that the interaction forces in the multi-stage NMPC remain low, unlike in the non-robust NMPC. Additionally, the probability of occurrence for Scenario 3 converges to 100% within a short time.

**Fig. 7.** Multi-stage NMPC joints angle (top graphs), angular velocity (middle graphs) and control inputs (bottom graphs) for 2Kg load

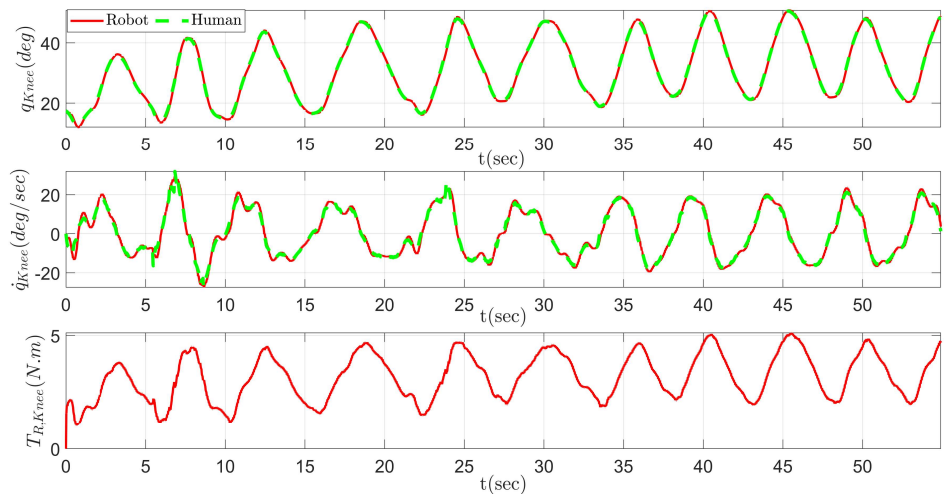

**Fig. 8.** NMPC and Multi-Stage NMPC tracking error of the angles (top graph), NMPC and Multi-Stage NMPC interaction forces (middle graph), and the probability of occurrence of the scenarios in Multi-Stage NMPC for 2Kg load (bottom graph).

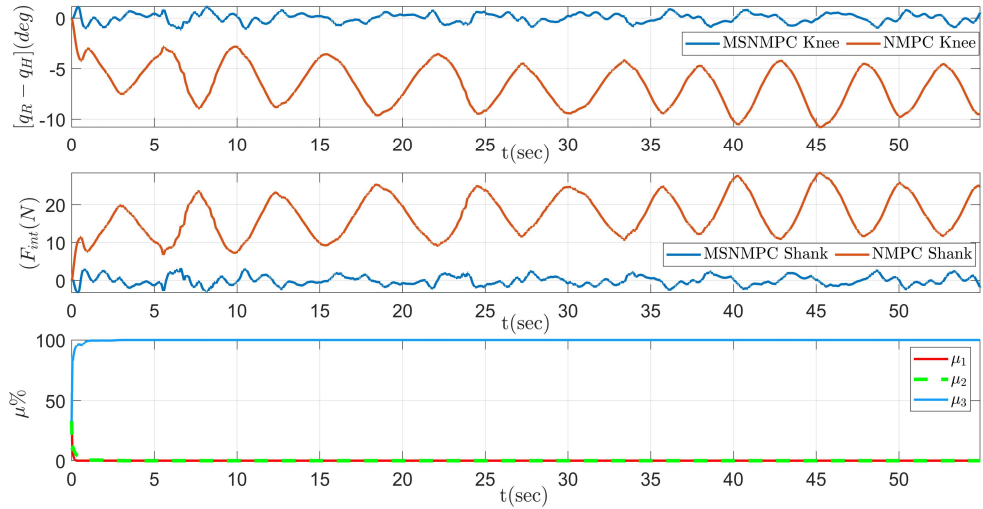

## 4.2. Experiments

The simulation results must be validated experimentally to evaluate the controller's functionality. For this purpose, a 1-DOF motorized exoskeleton robot is employed. The control system hardware consists of a computer, with incremental and absolute encoders measuring the joint angle and velocity, respectively. A force sensor (calibrated for the range of −70 N to +70 N) is integrated into the platform. All control and model parameters match those used in the simulation. The experimental setup is shown in Fig. 9.

**Fig. 9.** Experimental Exoskeleton Robot Platform.

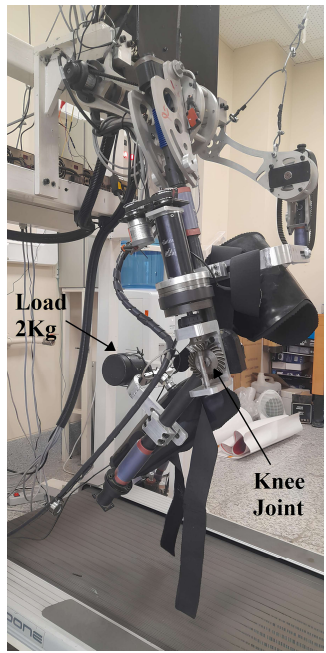

In experimental tests, the human operator defines the movement trajectory, which may introduce variability when comparing multi-stage NMPC with non-robust NMPC. Since the user's leg motion might differ between tests, this could affect the performance evaluation. To mitigate this issue, the robot's joint angle is monitored in real-time to ensure consistent trajectory tracking by human across both tests. Fig. 10 presents the knee joint angle and velocity for both tests, demonstrating that the executed trajectories remain closely matched. The experimental results are presented in Fig. 11. As shown, the non-robust controller exhibits an offset in interaction forces due to its inability to calculate appropriate compensation torques for gravitational effects. Additionally, it displays large-amplitude fluctuations because of its non-robust nature. In contrast, the robust controller maintains low interaction forces effectively. Notably, Scenario 3 demonstrates significantly higher occurrence probability compared to other scenarios, converging to 100% within 17 seconds of operation. Furthermore, the experimental results closely match the simulation outcomes, validating the controller's performance.

**Fig. 10.** Knee joint angle and angular velocity during experimental comparison of multi-stage NMPC versus non-robust NMPC controllers

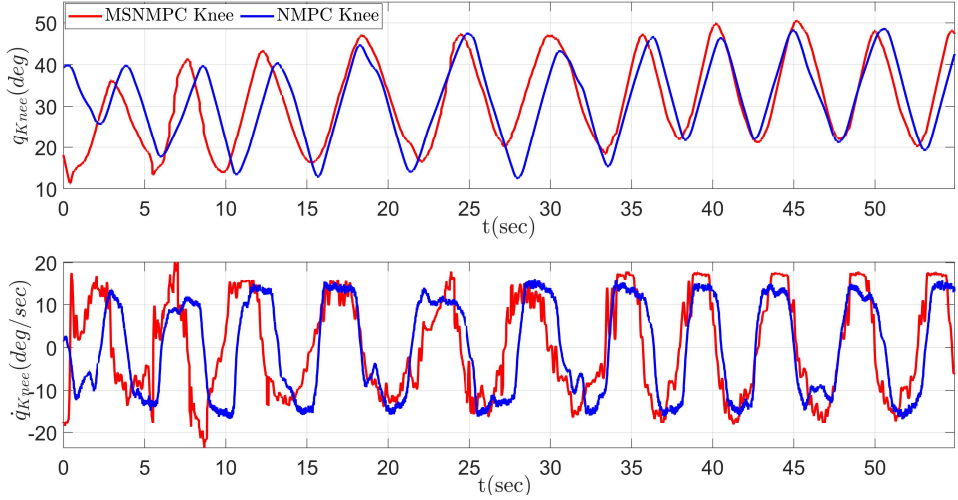

**Fig. 11.** NMPC and Multi-Stage NMPC interaction forces (top graph), Multi-Stage NMPC knee joint's torque (middle graph), and the probability of occurrence of the scenarios in Multi-Stage NMPC for 2Kg uncertainty (bottom graph).

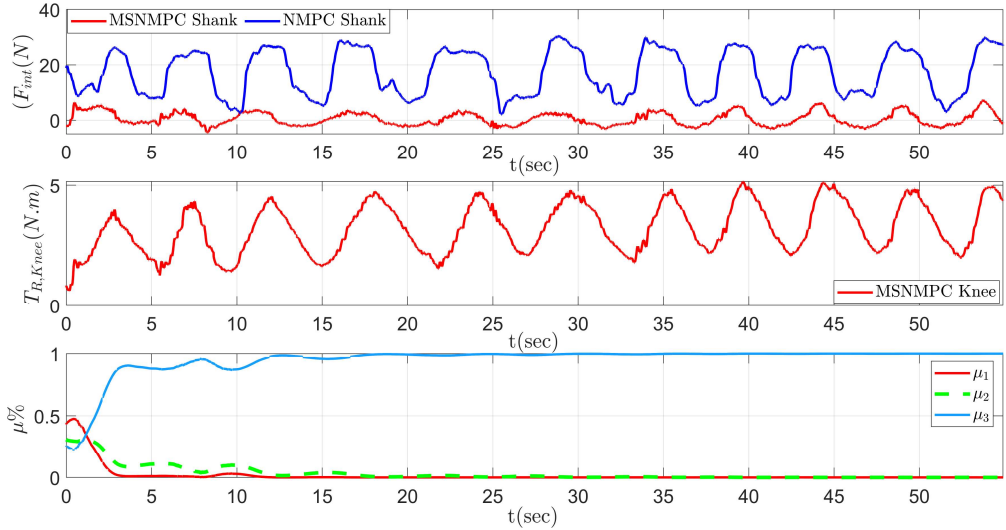

## 5. Conclusion

In this paper, multi-stage NMPC was used to track human motion and minimize human-robot interaction forces during the swing phase of normal gait using a 2-DOF exoskeleton robot. The system uncertainty was modeled as: (1) an unknown payload (0-2 kg) attached to the midpoint of the shank link, and (2) external disturbances, unmodeled dynamic terms, and measurement noise. Both simulation and experimental results conclusively demonstrate that multi-stage NMPC outperforms non-robust NMPC in controlling the exoskeleton robot, with smaller tracking errors and interaction forces observed in the presence of various uncertainties. For the case where the magnitude of the unknown mass was 2 kg, the RMS value of thigh and shank interaction forces under non-robust NMPC control were measured at 17.47× and 4.48× higher, respectively, compared to multi-stage NMPC. These findings suggest that multi-stage NMPC is a robust and effective control strategy for exoskeleton robots under uncertain conditions. For future work, the proposed controller's performance will be evaluated across additional human activities, including (e.g., stair ascent, sit-to-stand transitions) under varying dynamic conditions.

## Competing interests

The authors declare there are no competing interests.

## Author contribution

Conceptualization: AA, Data curation: AA, Formal analysis: AA, Funding acquisition: GV, Investigation: AA, Methodology: AA, Project administration: GV, AA, Software: AA, Resources: AA, Supervision: GV, Validation: AA, Visualization: AA, Writing – original draft: AA, Writing – review & editing: AA

## Funding

The authors declare no specific funding for this work.

## Data availability

Data generated or analyzed during this study are available from the corresponding author upon reasonable.